\documentclass[letterpaper]{article} 
\usepackage{aaai2027}  
\usepackage[hyphens]{url}  
\usepackage{graphicx} 
\usepackage{natbib}  
\usepackage{caption} 
\usepackage{algorithm}
\usepackage{algorithmic}

\usepackage{newfloat}
\usepackage{listings}
\DeclareCaptionStyle{ruled}{labelfont=normalfont,labelsep=colon,strut=off} 
\floatstyle{ruled}
\newfloat{listing}{tb}{lst}{}
\floatname{listing}{Listing}

\usepackage{booktabs}
\usepackage{bbding}

\nocopyright
\usepackage{amsmath}
\usepackage{amssymb}
\usepackage[table]{xcolor}
\usepackage{xspace}

\newcommand{\method}{AMBER\xspace}
\newcommand{\avlatent}{\mathbf{z}}
\newcommand{\vmem}{\mathbf{m}^{v}}
\newcommand{\amem}{\mathbf{m}^{a}}

\renewcommand{\method}{TaoMate\xspace}

\title{\method: Anchor-Guided Memory Bridging Evolving and Reference States for Real-Time Audio-Video Digital Human Generation}
\author{
    Qijun Gan\textsuperscript{\rm 1}\footnote{Project leader.}, 
    Chenwei Zhang\textsuperscript{\rm 1},
    Meiguang Jin\textsuperscript{\rm 1}\corresponding,
    Junfeng Ma\textsuperscript{\rm 1}, Qiu Shen\textsuperscript{\rm 2}
}
\affiliations{
    \textsuperscript{\rm 1}Taobao \& Tmall Group of Alibaba ~~
    \textsuperscript{\rm 2}Nanjing University\\

    \{ganqijun.gqj, meiguang.jmg\}@taobao.com
}

\begin{document}

\maketitle
\begin{abstract}
Real-time long-form digital-human generation relies on causal models to extend audio-visual content while preserving subject appearance and audio-video synchronization across successive segments.
A bounded cache retains local motion and phonetic context but discards older evidence, whereas attending to the complete generated history is computationally expensive and can propagate accumulated errors.
We present \method, an anchor-guided persistent-memory framework for few-step joint audio-video generation.
The framework preserves an immutable visual anchor, compresses completed video and audio blocks into fixed-capacity dynamic states, and retrieves those states through modality-specific residual attention without extending the active cache.
A reference-aware modulation method additionally conditions video features on dynamic and anchor appearance statistics.
Anchor-preserving causal-context distillation varies rollout horizon, prefix provenance, and cache-history reliability while keeping the immutable visual anchor unperturbed.
By separating persistent memory from stage-local denoising dependencies, \method further admits stage-parallel execution across blocks, accelerating autoregressive inference without pipeline-specific retraining.
We evaluate long-form video continuations with appearance, temporal, synchronization, facial, and speech diagnostics.
Results show that \method preserves stable appearance across prompt-conditioned segments and strong audio-visual synchronization under autoregressive generation. Our project page is \url{https://taoliveaigc.github.io/TaoMate}. 
\end{abstract}

\section{Introduction}
\begin{figure*}[t]
    \centering
    \includegraphics[width=0.95\textwidth]{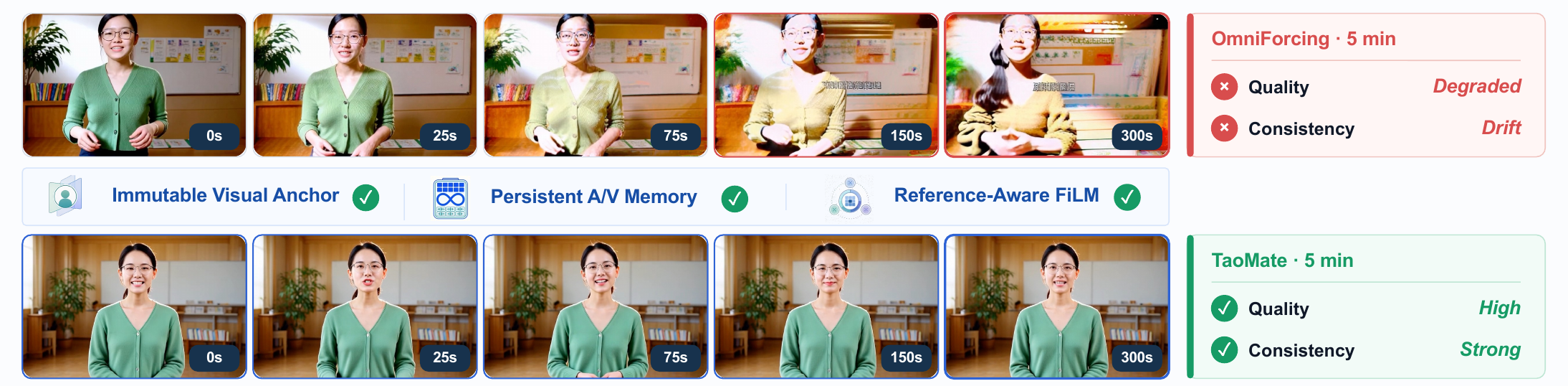}
    \vspace{-0.05in}
    \caption{
    \textbf{\method remains stable over 1{,}000 seconds.}
    Unlike OmniForcing's visual degradation, identity drift, and color shift, \method preserves consistency through visual anchoring, persistent A/V memory, and reference-aware FiLM.
    }
    \vspace{-0.15in}
    \label{fig:teaser}
\end{figure*}
Joint audio-video diffusion models have substantially advanced digital-human generation by modeling speech, facial motion, and body dynamics within a unified generative process~\cite{ltx2_2026,javisditpp2026,ovi2025,mova2026}.
Recent causal distillation methods further convert short-clip diffusion priors into few-step autoregressive generators~\cite{selfforcing2025,causalforcing2026,liveavatar2025,mutualforcing2026}, making continuous generation increasingly practical.
Long-form autoregressive generation, however, introduces a different requirement: an ordered sequence of prompts must drive evolving actions and speech while preserving the subject, scene, and acoustic behavior established earlier in the video.

Despite this progress, two interrelated challenges remain.
\textbf{i) Long-term audio-visual consistency.}
A bounded key-value (KV) cache retains the recent motion and phonetic context needed for local continuity, but earlier subject, scene, and acoustic evidence eventually leaves the cache.
Retaining the complete generated history instead causes attention and storage to grow with duration and repeatedly conditions the model on old, imperfect predictions.
Under recursive continuation, small errors in facial structure, color, scene layout, or articulation can therefore propagate across prompt transitions.
Moreover, causal distillation alone does not eliminate this accumulation because teacher trajectories provide clean bidirectional context, whereas deployment must consume the student's own causal outputs.
\textbf{ii) Efficient autoregressive inference.}
Each temporal block must traverse multiple denoising stages in a large joint audio-video model.
Conventional block-wise execution completes every stage for one block before advancing the next, leaving causally ready computation across blocks unexploited.

To address these challenges, we propose \method, an anchor-guided persistent-memory framework for long-form causal audio-video generation.
The framework decomposes temporal conditioning into a bounded, position-bearing active context and a fixed-capacity persistent memory.
The active context preserves fine-grained local motion and phonetic continuity, whereas persistent memory retains duration-independent visual and acoustic evidence after the corresponding tokens leave the cache.
This decomposition separates short-range positional continuity from long-range evidence while keeping the conditioning cost independent of generated duration.

Our framework promotes stable recursive generation through two complementary mechanisms.
First, persistent memory separates immutable visual and audio references from adaptive audio-video states.
The fixed references preserve the initial subject, scene, and acoustic characteristics, while adaptive states integrate detached representations of completed blocks; anchor agreement further regulates visual updates to suppress unreliable observations.
We further separate normalized content representations from channel-level and low-frequency appearance statistics: modality-specific memory attention retrieves structural and acoustic history, while reference-aware FiLM~\cite{film2018} reconciles dynamic appearance with the anchor.
Second, rollout-aligned causal distillation exposes the student to deployment-like causal contexts.
It combines mixed rollout horizons, student-generated prefixes, and selective perturbation of non-anchor history to approximate the context distribution encountered during recursive inference.
At inference, the separation between persistent memory and stage-local KV dependencies further enables stage-parallel execution.
Successive denoising stages are assigned to different devices, allowing causally ready blocks to advance concurrently after pipeline fill.
Only completed clean blocks update persistent memory, preserving the learned causal transition without pipeline-specific retraining.
Experiments on long-form Mandarin and English continuations show that \method maintains stable appearance and strong audio-video synchronization across prompt transitions.
On three GPUs, the stage-parallel implementation reaches $35$ output FPS, exceeding the 24-fps playback rate and enabling real-time long-form generation.

Our main contributions are as follows:
\begin{itemize}
    \item We introduce anchor-guided persistent multimodal memory that couples an immutable visual anchor with adaptive audio-video states beyond the active KV cache.
    \item We formulate rollout-aligned causal distillation using mixed rollout horizons and perturbed student histories to reduce exposure bias in recursive generation.
    \item We enable training-free stage-parallel inference across causally ready blocks, achieving real-time generation.
\end{itemize}

\section{Related Work}

\paragraph{Video and audio-video generation.}
Modern video systems combine diffusion, latent modeling, guidance, and transformer backbones~\cite{ddpm2020,ddim2021,cfg2022,ldm2022,dit2023,videoldm2023,stablevideodiffusion2023,ltxvideo2024}.
Joint generators model audio and video with asymmetric streams, twin backbones, or experts~\cite{ltx2_2026,javisditpp2026,ovi2025,mova2026}.
Portrait systems instead emphasize lip synchronization, motion, and appearance under specialized controls~\cite{wav2lip2020,makeittalk2020,audio2head2021,sadtalker2023,geneface2023,emo2024,hallo2024,liveportrait2024}.
These methods establish short-horizon generative priors, while long-form causal generation motivates mechanisms for retaining information across context refreshes.

\paragraph{Streaming generation and distillation.}
In streaming generation, exposure bias arises because training can condition on teacher-derived history whereas inference must consume model-generated history.
Self-Forcing and Causal Forcing train causal students on self-generated contexts~\cite{selfforcing2025,causalforcing2026}.
Mutual Forcing instead couples few-step and multi-step modes in a weight-shared autoregressive model, using the few-step mode to construct history for self-distillation~\cite{mutualforcing2026}.
Live Avatar combines causal distillation with long-horizon streaming strategies~\cite{liveavatar2025}.
AvatarForcing performs one-step sliding-window denoising with a RoPE-reindexed style anchor, recent clean temporal anchors, and two-stage distribution-matching distillation~\cite{avatarforcing2026}.
Few-step generation has also been studied through progressive, consistency, latent-consistency, and distribution-matching distillation~\cite{progressive_distill2022,consistency_models2023,latent_consistency2023,dmd2023,dmd2_2024}.
PCM decomposes a probability-flow trajectory into adjacent-stage consistency problems~\cite{pcm2024}.
Several of these methods primarily alter how a causal trajectory or its supervision is constructed.
Our training scheme couples self-generated history construction with cache-external memory and keeps the immutable visual anchor unperturbed when perturbing later prefixes.

\paragraph{Position-bearing history retention.}
Long-context models employ recurrence, compression, retrieval, sinks, heavy-hitter retention, and blockwise attention~\cite{transformerxl2019,compressive_transformer2019,memorizing_transformer2022,streamingllm2023,h2o2023,ringattention2023}.
For video, TetherCache recalls historical K/V and aligns selected entries to trusted cache statistics, while Pyramid and Sparse Forcing organize retained position-bearing context~\cite{tethercache2026,pyramidforcing2026,sparseforcing2026}.
OmniMem sparsely retrieves query-relevant blocks from the full historical K/V cache, whereas FadeMem consolidates older K/V blocks into a distance-aware temporal hierarchy~\cite{omnimem2026,fademem2026}.
Anchor Forcing instead stores anchor K/V for prompt-switch re-caching and uses region-specific RoPE origins to control positional drift~\cite{anchorforcing2026}.
MaineCoon combines agentic prompt planning with inference-time cache management, while AvatarForcing reindexes a style anchor and retains recently emitted clean blocks~\cite{mainecoon2026,avatarforcing2026}.
These approaches preserve, retrieve, consolidate, or reposition elements of position-bearing history.

\section{Method}
\label{sec:method}

\subsection{Problem Formulation}

We consider autoregressive audio-video generation from a temporally ordered sequence of text prompts.
The output unfolds causally as a variable-length sequence of synchronized video-audio latent blocks, where each prompt conditions a contiguous temporal segment.
Let $\avlatent_t=(\avlatent_t^v,\avlatent_t^a)$ denote the $t$-th block and let $e_t$ denote its associated text condition.
A block is the atomic unit of generation, while consecutive blocks sharing the same condition form a prompt-level output segment.
This setting imposes two distinct requirements on temporal conditioning.
Fine-grained motion, pose, and phonetic continuity depend on recent latent tokens with explicit temporal positions, whereas subject, scene, and acoustic evidence can remain relevant after those tokens leave a finite attention cache.
Retaining the complete generated history would make attention and storage grow with duration and repeatedly expose the generator to old prediction errors; retaining only recent tokens would instead discard long-range evidence.
We therefore decompose history into a bounded active context $\mathcal{C}_{t-1}$ and a fixed-capacity persistent memory $\mathcal{M}_{t-1}$.
The former preserves position-bearing recent tokens for local continuity, while the latter compresses detached historical evidence for duration-independent long-range conditioning.
The resulting causal factorization is
\begin{equation}
\begin{aligned}
    \avlatent_t &\sim p_\theta(\avlatent_t\mid
    e_t,\mathcal{C}_{t-1},\mathcal{M}_{t-1}),\\
    \mathcal{M}_t &= U(\mathcal{M}_{t-1},
    \operatorname{sg}(\avlatent_t)).
\end{aligned}
\label{eq:causal_factorization}
\end{equation}
Here $\operatorname{sg}$ denotes stop-gradient, which avoids backpropagation through an unbounded chain of memory updates.
The update is applied only after $\avlatent_t$ has been generated, so the condition for block $t$ contains information only from preceding blocks.
Because both $\mathcal{C}_{t-1}$ and $\mathcal{M}_{t-1}$ have bounded capacity, their storage and conditioning cost do not grow with the duration already generated.

\subsection{Anchor-Guided Persistent Memory}
\label{sec:persistent_memory}

Long-horizon generation requires memory to preserve information with different temporal semantics. Subject appearance and acoustic characteristics should remain stable under recursive generation, whereas motion, scene evolution, and phonetic context must adapt to newly generated content. Encoding these roles in a single recurrent representation couples retention and adaptation: rapid updates can overwrite identity-defining evidence, while conservative updates limit responsiveness to evolving context. We therefore factor persistent memory into fixed references and adaptive modality-specific states:
\begin{equation}
    \mathcal{M}_t=(\mathbf{a}^{v},\mathbf{a}^{a},
    \vmem_t,\amem_t,\boldsymbol\kappa_t,\boldsymbol\pi_t).
    \label{eq:persistent_state}
\end{equation}
The visual anchor $\mathbf{a}^{v}$ and audio reference $\mathbf{a}^{a}$ retain canonical evidence from the initial block. The content memories $\vmem_t$ and $\amem_t$ aggregate evolving spatial and acoustic structure, while $\boldsymbol\kappa_t$ and $\boldsymbol\pi_t$ preserve absolute channel and low-frequency appearance statistics that are separated from normalized visual content. This factorization allows each information type to follow an update rule and conditioning pathway matched to its temporal role.

\begin{figure*}[t]
\centering
\includegraphics[width=0.95\textwidth]{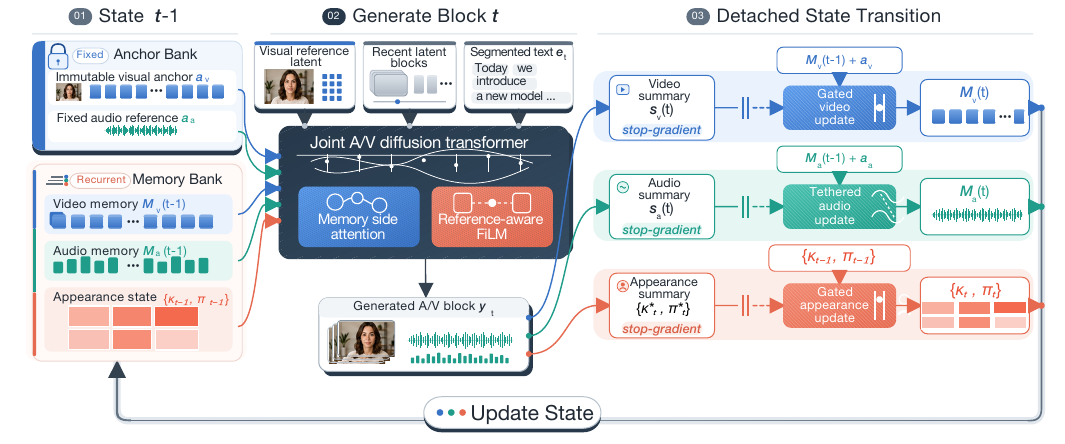}
\vspace{-0.1in}
\caption{\textbf{\method Pipeline.} The pipeline generates synchronized long-form video and audio causally with anchor-guided persistent memory. A bounded context provides local motion and phonetic cues, while a fixed visual anchor and adaptive modality states supply long-range conditioning through side attention and reference-aware FiLM. }
\vspace{-0.15in}
\label{fig:pipeline}
\end{figure*}

All block-derived memory states are detached and have fixed cardinality, which prevents an unbounded optimization graph and keeps memory cost independent of generated duration. The compressed visual anchor is distinct from the uncompressed, position-bearing initial visual latent retained in the active context $\mathcal{C}_t$: the former provides compact long-range reference evidence, whereas the latter remains an explicit token-level condition for causal attention.
Figure~\ref{fig:pipeline} summarizes the state evolution.

\paragraph{Factorized block representations.}
The fixed references are instantiated once from the initial block, whereas the adaptive memory components must incorporate evidence from each completed block. To update them without conflating relative content structure with drift-prone absolute appearance, we decompose each block into normalized modality content representations and unnormalized visual appearance representations:
\begin{equation}
\begin{aligned}
    \mathbf{s}_t^v&=P_v[\operatorname{Norm}_{HW}(\bar{\mathbf z}_t^v)],
    &\bar{\mathbf z}_t^v&=F_v^{-1}\textstyle\sum_f\avlatent_{t,f}^v,\\
    \mathbf{s}_t^a&=P_a[\operatorname{Norm}_{F_a}(\avlatent_t^a)],
    &\boldsymbol\kappa_t^*&=K(\avlatent_t^v),\quad
    \boldsymbol\pi_t^*=P_c(\bar{\mathbf z}_t^v).
\end{aligned}
\label{eq:state_candidates}
\end{equation}
Here $\mathbf{s}_t^v$ and $\mathbf{s}_t^a$ denote the fixed-cardinality video and audio content representations extracted from block $t$; they serve as observations for updating $\vmem_t$ and $\amem_t$, rather than as persistent states themselves. $F_v$ and $F_a$ are the temporal lengths of the video and audio latents, respectively, and $f$ indexes video latent frames. $\operatorname{Norm}_{HW}$ standardizes each video channel over its spatial dimensions, whereas $\operatorname{Norm}_{F_a}$ standardizes each audio channel over the temporal dimension. $P_v$, $P_a$, and $P_c$ are fixed-cardinality projections, and $K$ extracts channel-wise mean and scale. The superscript $*$ denotes a current-block candidate before it is incorporated into persistent memory. The normalized content representations retain relative spatial and acoustic structure, whereas the appearance observations $\boldsymbol\kappa_t^*$ and $\boldsymbol\pi_t^*$ preserve color, contrast, and coarse spatial statistics for long-range calibration.

\paragraph{Anchor-constrained memory evolution.}
Block-wise representations extracted from generated history are imperfect observations. Incorporating them at a fixed rate allows transient prediction errors to accumulate in future conditions, whereas freezing the memory prevents legitimate motion and speech evolution. We therefore formulate memory evolution as an anchor-regularized online estimate:
\begin{equation}
    \mathcal{E}(\mathbf{m},\mathbf{a},\mathbf{s};\beta,\lambda)
    =(1-\lambda)[(1-\beta)\mathbf{m}+\beta\mathbf{s}]
    +\lambda\mathbf{a}.
    \label{eq:tethered_ema}
\end{equation}
The coefficients of the previous estimate, current observation, and fixed reference are $(1-\lambda)(1-\beta)$, $(1-\lambda)\beta$, and $\lambda$, respectively. They are nonnegative and sum to one, making $\mathcal{E}$ their weighted Euclidean barycenter. Thus $\beta$ controls evidence assimilation, while $\lambda$ imposes a persistent reference prior. For constant rates, the coefficient of an observation from $k$ updates earlier is $(1-\lambda)\beta[(1-\lambda)(1-\beta)]^k$; historical evidence and isolated errors therefore decay geometrically, while the anchor is reintroduced at every update.

The reliability of a video observation depends on whether it preserves the structural evidence established by the reference. Because $\mathbf{s}_t^v$ is spatially normalized, its agreement with the visual anchor is insensitive to first-order channel shifts. We convert this agreement into a smooth observation-confidence gate:
\begin{equation}
\begin{aligned}
    \rho_t&=B^{-1}\sum_b
    \cos(\operatorname{vec}(\mathbf{s}_{t,b}^v),
    \operatorname{vec}(\mathbf{a}^{v}_b)),\\
    g_t&=g_{\min}+(1-g_{\min})
    \sigma((\rho_t-\tau)/T_g),\\
    \vmem_t&=\mathcal{E}(\vmem_{t-1},\mathbf{a}^{v},
    \mathbf{s}_t^v;\beta_v g_t,\lambda_v),\\
    \amem_t&=\mathcal{E}(\amem_{t-1},\mathbf{a}^{a},
    \mathbf{s}_t^a;\beta_a,\lambda_a).
\end{aligned}
\label{eq:memory_update}
\end{equation}
Here $B$ denotes the mini-batch size and $b$ indexes samples within the mini-batch. $\operatorname{vec}$ flattens the token and channel dimensions, $\sigma(\cdot)$ is the sigmoid function, $\tau$ is the agreement threshold, and $T_g$ is the gate temperature.
Low agreement reduces the effective video update rate, limiting the influence of structurally inconsistent generations. The nonzero gate floor avoids irreversible freezing when valid pose or scene changes reduce similarity. Audio uses an ungated update because rapid phonetic variation is expected rather than evidence of identity drift, while its anchor tether still preserves long-term acoustic characteristics.

Absolute appearance requires stronger protection because a channel-level error can directly propagate as visible color or contrast drift. We therefore apply the same soft confidence gate to $\boldsymbol\kappa_t$ and $\boldsymbol\pi_t$, together with a hard rejection rule based on their normalized deviation from anchor statistics. In-range observations support gradual appearance adaptation, whereas outliers leave the previous appearance memory unchanged. The resulting memory is a robust, fixed-capacity estimator of long-range evidence rather than a lossless archive of generated blocks.

\subsection{Memory Conditioning}
\label{sec:memory_conditioning}

Persistent memory should influence generation without being converted back into additional position-bearing history. Directly concatenating all memory components with the active sequence would blur the distinction between temporally ordered local context and position-free long-range evidence. It would also force token-valued content and global appearance statistics through the same retrieval mechanism, despite their different roles in generation. We therefore use structure-matched conditioning: content memory supports query-dependent retrieval, whereas appearance statistics act coherently across video tokens through feature modulation.
For modality $m\in\{v,a\}$, let $\mathbf{r}_{t-1}^v=[\mathbf{a}^{v};\vmem_{t-1}]$, $\mathbf{r}_{t-1}^a=\amem_{t-1}$, and $\mathbf{u}_{t-1}^m=E_m(\mathbf{r}_{t-1}^m)$. Selected transformer layers then apply
\begin{equation}
\begin{aligned}
    \widetilde h_\ell^m&=h_\ell^m+
    \operatorname{MemAttn}_\ell^m(
    \operatorname{RMSNorm}(h_\ell^m),\mathbf{u}_{t-1}^m),\\
    \widehat h_\ell^v&=\widetilde h_\ell^v+
    \boldsymbol\gamma_\ell(\mathbf{c}_{t-1})\odot
    \operatorname{RMSNorm}(\widetilde h_\ell^v)
    +\mathbf{b}_\ell(\mathbf{c}_{t-1}),
\end{aligned}
\label{eq:memory_conditioning}
\end{equation}
where $\mathbf{c}_{t-1}$ concatenates dynamic and anchor channel statistics and the second line is a reference-aware FiLM residual~\cite{film2018}.
The low-frequency appearance state $\boldsymbol\pi_{t-1}$ follows a complementary parameter-free readout that calibrates each completed video latent before it is committed as future context and used to update memory.
Memory K/V remain outside the active cache, so fixed token counts keep the additional cost independent of generated duration.
Both pathways are residual and initialized as identity mappings; architectural details are deferred to the supplementary material.

\subsection{Causal-Context Distillation and Generation}
\label{sec:rollout_distillation}

Autoregressive distillation introduces a context-distribution gap: teacher trajectories provide clean bidirectional context, whereas the deployed student must continue from its own imperfect causal history.
We reduce this gap by distilling on causal self-rollouts that span multiple temporal horizons and draw prefixes from both clean trajectories and detached student generations.
Before intermediate non-anchor blocks are reused as causal context, their latent representations are perturbed along the forward flow-matching path, while the persistent visual anchor remains clean.
This asymmetric treatment discourages the student from treating recent self-generated context as perfectly reliable and preserves the anchor as a stable source of appearance evidence.
Together, these variations expose the student to accumulated generation errors without weakening the persistent identity signal.

On these causal contexts, distribution-matching distillation transfers the guided teacher distribution over the sampled rollout~\cite{dmd2023,dmd2_2024}.
We complement this global objective with video PCM~\cite{pcm2024}, which promotes consistency between neighboring denoising transitions in latent and appearance-sensitive feature spaces.
Persistent-memory updates are detached from the optimization graph, allowing extended causal rollouts without retaining gradients through the complete history.
The full objectives and sampling schedules are provided in the supplementary material.

\subsection{Memory-Aware Stage-Parallel Inference}
\label{sec:stage_parallel_inference}

\begin{figure}[t]
    \centering
    \includegraphics[width=\columnwidth]{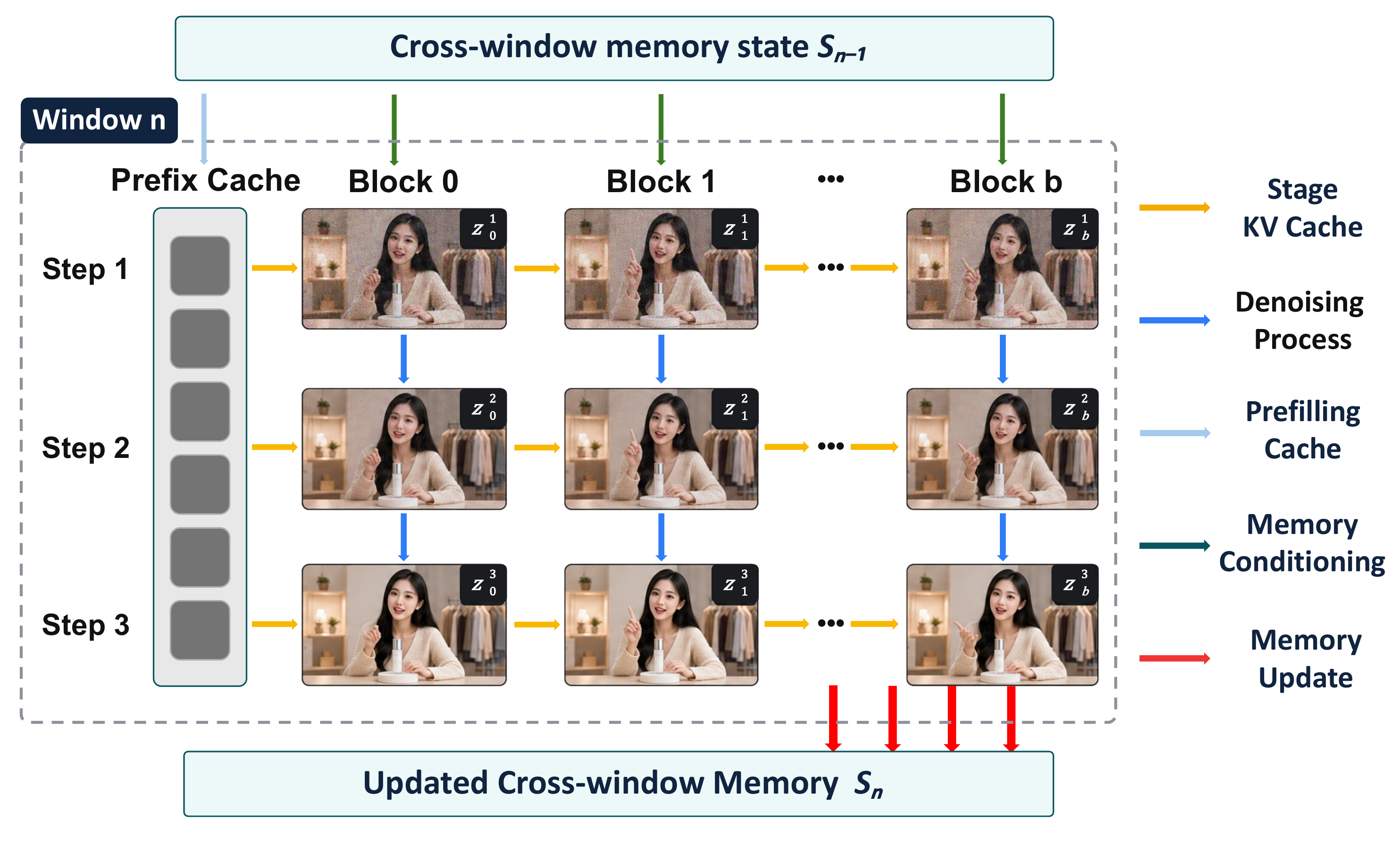}
    \vspace{-0.20in}
    \caption{
        Memory-aware stage-parallel inference. Each stage pre-fills its KV cache from the bounded clean prefix $\mathcal{P}_{n-1}$ and reads the persistent memory $\mathcal{M}_{n-1}$. Blocks advance along an anti-diagonal pipeline, and only final clean outputs update cross-window conditioning.
    }
    \vspace{-0.20in}
    \label{fig:streaming_pipeline_denoising}
\end{figure}

Generating a long video in a single pass would cause the attention context
and its position-bearing KV cache to grow continuously.
We therefore divide the video into a sequence of bounded windows and propagate
the cross-window conditioning pair
$\mathcal{S}_{n-1}=(\mathcal{P}_{n-1},\mathcal{M}_{n-1})$.
The clean prefix $\mathcal{P}_{n-1}$ provides explicit latents for short-range continuity, whereas the fixed-capacity persistent memory $\mathcal{M}_{n-1}$ provides learned representations for long-range identity and appearance consistency.
In our configuration, the prefix consists of the initial reference
sink and up to recent blocks.

Within each window, every pipeline rank is assigned one denoising stage and
maintains an independent video--audio KV cache.
Before denoising starts, each stage initializes its cache by pre-filling the
same explicit clean latents from $\mathcal{P}_{n-1}$, while the persistent
memory $\mathcal{M}_{n-1}$ conditions both the prefix prefill and all
subsequent denoising forwards.
This cross-window prefix is prefilled only once per window. At the window boundary, only final-stage clean blocks construct $\mathcal{P}_n$ and update
$\mathcal{M}_n$ for window $n+1$.
Each stage discards its previous window cache and pre-fills a new cache from
initial reference sink and the most recent clean blocks using positions local to the new window.

During window generation, a block completed at stage $k$ is committed only
to the cache of stage $k$, as indicated by the yellow arrows in
Fig.~\ref{fig:streaming_pipeline_denoising}.
Consequently, when stage $k$ processes block $b$, its cache contains the
bounded clean prefix and all preceding blocks
$z_0^k,\ldots,z_{b-1}^k$ committed at the same denoising stage.
Cleaner outputs from downstream stages are never written back into an
upstream cache.
This stage-local cache organization bounds the position-bearing context while
preserving causal access to all earlier same-stage blocks. A task $(b,k)$ depends on the same block at the preceding stage,
$(b,k-1)$, and on the cache state produced by $(b-1,k)$, which
already contains all earlier same-stage blocks.
Tasks satisfying $b+k=d$ therefore form an anti-diagonal wavefront.
After pipeline fill, different blocks are processed simultaneously at
different denoising stages, and the final stage continuously retires clean
blocks.
This substantially improves inference FPS over serial block-wise denoising,
enabling real-time autoregressive long-video generation. Unlike HiAR~\cite{zou2026hiar}, our method achieves
quality-preserving acceleration without pipeline-specific retraining.

\section{Experiments}
\label{sec:experiments}

\begin{table*}[t]
\caption{Comparison with representative speech-to-video (S2V) methods and other baselines on the long-form benchmark. Best and second-best results are shown in bold and underlined, respectively.}
\vspace{-0.1in}
\label{tab:external_results}
\centering
\setlength{\tabcolsep}{2.4pt}
\renewcommand{\arraystretch}{1.08}
\begin{tabular}{lcccccccc}
\toprule
Method & Params & \shortstack{LipSync$\uparrow$} & \shortstack{Desync$\downarrow$} & \shortstack{Align.$\uparrow$} & \shortstack{Express.$\uparrow$} & \shortstack{Long Cons.$\uparrow$} & \shortstack{Color $\Delta\downarrow$} & \shortstack{DiT FPS$\uparrow$} \\
\midrule
OmniAvatar~\cite{omniavatar2025} & 1.3B & 4.794 & -- & -- & -- & 0.6467 & 0.01566 & 0.948 \\
StableAvatar~\cite{stableavatar2025} & 1.3B & 3.246 & -- & -- & -- & 0.9242 & 0.00895 & 0.352 \\
Wan2.2-S2V-14B~\cite{wans2v2025} & 14B & 4.898 & -- & -- & -- & 0.8413 & 0.00814 & 0.079 \\
LiveAvatar~\cite{liveavatar2025} & 14B & \underline{5.396} & -- & -- & -- & \underline{0.9619} & \underline{0.00541} & 4.529 \\
Hallo3~\cite{hallo3_2025} & 5B & 4.855 & -- & -- & -- & 0.7947 & 0.01383 & 0.150 \\
LongLive-2.0~\cite{longlive2_2026} & 5B & -- & -- & -- & -- & 0.7221 & 0.01792 & \underline{10.47} \\
\midrule
LTX-2.3 ~\cite{ltx2_2026} & 22B & 2.349 & 0.050 & 4.35 & 3.60 & 0.3745 & 0.04387 & 0.262 \\
JavisDiT++~\cite{javisditpp2026} & 2.1B & 0.904 & 1.025 & \textbf{5.00} & 3.90 & 0.3545 & 0.07159 & 0.437 \\
MOVA~\cite{mova2026} & 32B/18B & 3.860 & 0.060 & \textbf{5.00} & 3.75 & 0.7129 & 0.02825 & 0.128 \\
OVI~\cite{ovi2025} & 11B & 3.493 & 0.135 & \underline{4.60} & 3.70 & 0.8252 & 0.01797 & 0.231 \\
\midrule
OmniForcing & 22B & 1.590 & \underline{0.045} & \textbf{5.00} & \underline{3.95} & 0.3404 & 0.02060 & 8.14 \\
\method & 22.1B & \textbf{5.918} & \textbf{0.000} & \textbf{5.00} & \textbf{4.00} & \textbf{0.9755} & \textbf{0.00260} & \textbf{16.32} \\
\bottomrule
\vspace{-0.2in}
\end{tabular}
\end{table*}

\begin{figure*}[t]
    \centering
    \includegraphics[width=0.95\textwidth]{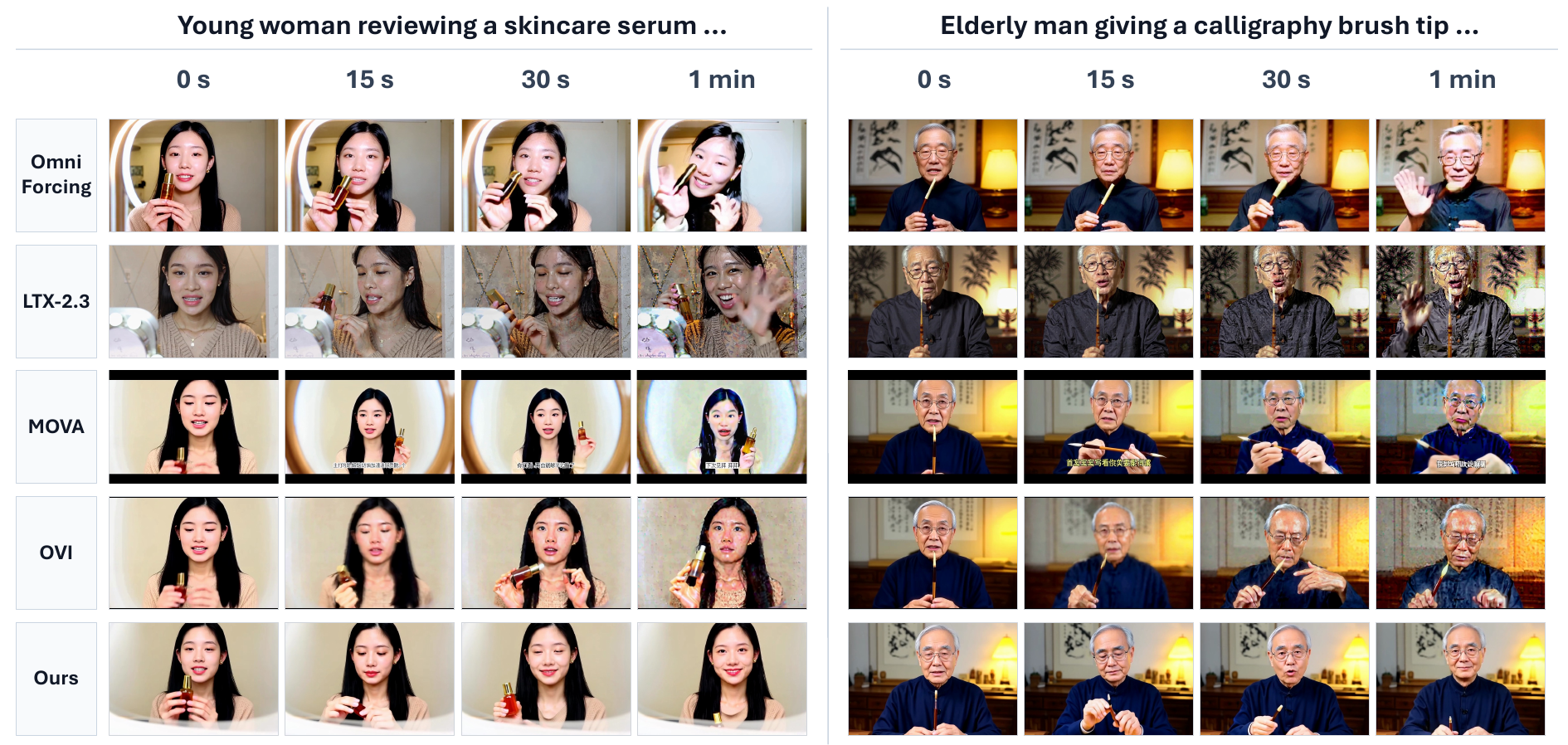}
    \vspace{-0.1in}
    \caption{
        \textbf{Qualitative comparison on one-minute video generation.}
        For two representative prompts, we show frames sampled at
        0, 15, 30, and 60 seconds (columns) from videos generated by different
        methods (rows).
    }
    \vspace{-0.15in}
    \label{fig:visual_quality}
\end{figure*}

\subsection{Implementation Details}

\noindent\textbf{Long-form benchmark.}
We evaluate single-presenter scenarios in both Mandarin and English.
Each scenario is represented by a temporally ordered prompt sequence in which every prompt specifies the action, transcript, subject appearance, scene, camera, and voice for one short segment.
The subject, scene, camera, and voice descriptions remain consistent across the sequence, while the action and spoken content evolve.
Every internal output contains 60 seconds of video at $768\times512$ and 24 fps.

\noindent\textbf{Training.}
The causal generator is initialized from a distilled LTX-2.3 checkpoint and optimized against a frozen bidirectional teacher using three denoising steps per block.
The teacher uses classifier-free guidance scales of $3.0$ for video and $5.0$ for audio.
The generator and critic models use learning rates of $10^{-5}$ and $2\times10^{-6}$, respectively, and the generator EMA decay is $0.99$.
Residual memory attention is inserted every four transformer layers, giving 12 memory-enabled layers in the 48-layer backbone.
The video branch receives 48 tokens formed by the visual anchor and dynamic video state, while the audio branch receives 64 dynamic-state tokens; together with the modality encoders and video FiLM, these branches add approximately $97.8$M trainable parameters.

Training samples rollout horizons of five and ten blocks with probabilities $0.8$ and $0.2$ and substitutes detached student-generated prefixes with probability $0.1$.
Intermediate generated blocks reused as cache history are re-noised with $\sigma\sim\mathcal{U}(0,0.15)$, while the persistent visual anchor is initialized from the unperturbed reference latent.
The adaptive formulation uses $(\beta_v,\lambda_v)=(0.15,0.20)$ and $(\beta_a,\lambda_a)=(0.10,0.10)$ for video and audio state updates, with gate parameters $(\tau,T_g,g_{\min})=(0.05,0.10,0.10)$.
Appearance states use $(\beta_c,\lambda_c)=(0.03,0.60)$.
The PCM regularizer has weight $0.03$, begins at optimization step 400, and is evaluated every two optimization steps.

We train on 6,139 filtered synthetic trajectories generated by the bidirectional LTX-2.3 teacher~\cite{ltx2_2026}.
Each trajectory contains paired video and audio ODE states, with up to 241 video frames at $768\times512$ and 24 fps.
The audio latents are aligned to the duration of the corresponding video trajectory.

\noindent\textbf{Inference.}
The EMA student generator produces each one-minute sample by following its ordered prompt sequence autoregressively.
At a prompt transition, the active context retains the initial visual reference and at most two recent non-anchor blocks.
Only completed blocks update the persistent memory, and local RoPE coordinates restart for each prompt-conditioned segment.
The end-to-end single-device implementation requires $130.3\pm1.3$ seconds per 1,441-frame output on an NVIDIA RTX PRO 5000 72GB GPU, corresponding to $11.1$ generated frames per second.

\subsection{Comparison with State-of-the-Art Methods}
\label{sec:sota_comparison}

\noindent\textbf{Evaluation metrics.}
We adopt LipSync, Desync, Alignment, and Expressiveness from the official VABench evaluation protocol~\cite{vabench2026}.
For audio-conditioned video-only methods, LipSync is evaluated against the conditioning audio.

To capture long-form behavior beyond these audio-visual criteria, we use two complementary measures over fixed temporal intervals.
\emph{Color $\Delta$} measures local appearance discontinuity between neighboring intervals and is therefore sensitive to visible boundary-level color shifts.
\emph{Long Consistency} instead summarizes global appearance stability over the complete video, reflecting whether its color and brightness remain coherent over time.
Lower Color $\Delta$ and higher Long Consistency indicate better temporal stability; their exact definitions are provided in the supplementary material.
For the ablation study, \emph{Face Consistency} compares normalized MediaPipe facial geometry between early and late intervals~\cite{mediapipeface2019}; it measures face-shape stability rather than biometric identity.
\emph{DiT FPS} is the number of generated frames per denoising second and excludes data loading, VAE decoding, and post-processing. All methods are profiled on the same NVIDIA RTX PRO 5000 72GB GPU at $768\times512$ resolution.

\noindent\textbf{Baselines and evaluation setup.}
We compare with the joint audio-video generators LTX-2.3, JavisDiT++, OVI, and MOVA~\cite{ltx2_2026,javisditpp2026,ovi2025,mova2026}, together with an OmniForcing-aligned causal baseline.
We additionally include the audio-driven video generators OmniAvatar, StableAvatar, Wan2.2-S2V, LiveAvatar, and Hallo3, and the LongLive-2.0 text-to-video plus image-to-video pipeline~\cite{omniavatar2025,stableavatar2025,wans2v2025,liveavatar2025,hallo3_2025,longlive2_2026}.
LTX-2.3 follows unified chunked continuation, while JavisDiT++ and OVI generate native-length clips that are assembled according to the benchmark prompts.
All systems follow the same script content, but retain their native resolution, frame rate, and long-form construction procedure.

\noindent\textbf{Quantitative results.}
Table~\ref{tab:external_results} reports quantitative comparisons with representative S2V systems and joint audio-video baselines.
\method ranks first in LipSync, Desync, and Expressiveness and ties for the best Alignment, demonstrating accurate speech-motion coupling with expressive audio-video dynamics.
It also achieves the highest Long Consistency and lowest Color $\Delta$, indicating that repeated continuation preserves global appearance while suppressing local boundary shifts.
Relative to the strongest external result on each measure, these scores improve LipSync by $9.7\%$, raise Long Consistency by $0.0136$, and reduce Color $\Delta$ by $51.9\%$.
Joint gains in local and global stability show that suppressing boundary drift preserves coherent appearance evolution.
At $16.32$ DiT FPS, \method is $1.56\times$ faster than the next-highest measured throughput, showing that these stability gains remain compatible with efficient denoising.

\noindent\textbf{Qualitative results.}
Figure~\ref{fig:visual_quality} presents visual results from one-minute generations.
Across both sequences, \method preserves facial geometry, clothing appearance, and scene layout while allowing the prompted gestures to evolve.
In contrast, LTX-2.3 develops pronounced late-stage facial and background artifacts, while MOVA and OVI exhibit larger changes in facial appearance, framing, or scene composition.
These visual comparisons support the low Color $\Delta$ and high Long Consistency reported in Table~\ref{tab:external_results}.

\subsection{Inference Throughput}
\label{sec:multigpu_throughput}

\begin{table}[t]
\caption{Multi-GPU end-to-end inference throughput. Size denotes denoising-generator parameters.}
\label{tab:multigpu_fps}
\centering
\setlength{\tabcolsep}{3.5pt}
\renewcommand{\arraystretch}{1.08}
\vspace{-0.1in}
\begin{tabular}{lccc}
\toprule
Method & Size & GPUs & FPS $\uparrow$ \\
\midrule
LiveAvatar~\cite{liveavatar2025} & 14B & 5 & 23.0 \\
LongLive-2.0~\cite{longlive2_2026} & 5B & 4 & 16.0 \\
\method & 22.1B & 3 & 35.0 \\
\bottomrule
\vspace{-0.2in}
\end{tabular}
\end{table}

Table~\ref{tab:multigpu_fps} evaluates whether the compared systems sustain real-time generation under their reported multi-GPU deployments.
\method reaches 35.0 output FPS on three GPUs, exceeding its 24-fps playback rate by $45.8\%$ and the reported throughput of LiveAvatar and LongLive-2.0 by $1.52\times$ and $2.19\times$, respectively.
At this throughput, the complete 1,441-frame generation pipeline finishes in approximately $41.2$ seconds.
By restricting persistent-memory updates to fully denoised outputs, our stage-parallel schedule allows causally ready blocks to progress concurrently through different denoising stages, achieving real-time throughput.

\subsection{Ablation Study}

Relative to the memory-free baseline, fixed-capacity memory raises Long Consistency by $0.0039$ and lowers Color $\Delta$ by $16.8\%$, while Face Consistency changes by only $0.0001$.
This profile indicates that cross-window evidence mainly stabilizes global appearance and scene continuity, with facial geometry already near saturation.
Adding ungated FiLM, however, erodes these gains: Long and Face Consistency decrease by $0.0033$ and $0.0009$, respectively, while Color $\Delta$ increases by $9.8\%$.
Gating recurrent updates by anchor agreement instead raises the two consistency scores by $0.0056$ and $0.0015$ and reduces Color $\Delta$ by $20.5\%$, producing the best result on every metric.
This contrast shows that appearance modulation is effective when unreliable recurrent observations are suppressed before conditioning future blocks.

\begin{table}[H]
\caption{Ablation of persistent-memory components under the shared training protocol. Higher is better except Color $\Delta$. Best and second-best values are bold and underlined.}
\vspace{-0.1in}
\label{tab:system_variants}
\centering
\setlength{\tabcolsep}{1.7pt}
\renewcommand{\arraystretch}{1.08}
\begin{tabular}{cccccc}
\toprule
Memory & FiLM & Gate & \shortstack{Long Cons.$\uparrow$} & \shortstack{Color $\Delta\downarrow$} & \shortstack{Face Cons.$\uparrow$} \\
\midrule
-- & -- & -- & 0.9614 & 0.00380 & 0.9955 \\
$\checkmark$ & -- & -- & \underline{0.9653} & \underline{0.00316} & \underline{0.9956} \\
$\checkmark$ & $\checkmark$ & -- & 0.9620 & 0.00347 & 0.9947 \\
$\checkmark$ & $\checkmark$ & $\checkmark$ & \textbf{0.9676} & \textbf{0.00276} & \textbf{0.9962} \\
\bottomrule
\vspace{-0.05in}
\end{tabular}
\end{table}

\section{Conclusion}
\label{sec:conclusion}

We introduced \method, a framework for real-time long-form audio-video generation that separates a bounded, position-bearing active context from fixed-capacity persistent memory.
The memory combines immutable references with adaptive content and appearance states for stable long-range conditioning.
Rollout-aligned causal distillation further uses mixed rollout horizons, student-generated prefixes, and selective perturbation of non-anchor history to reduce the context gap between teacher-guided training and recursive inference.
Experiments and ablations demonstrate strong audio-video synchronization, improved cross-window appearance stability, and the importance of anchor-regulated memory updates.
This separation further enables stage-parallel inference, reaching $35$ output FPS on three GPUs.

\bibliography{aaai2027}

\end{document}